\documentclass{article}
\usepackage{ijcai24}
\usepackage{times}
\usepackage{soul}
\usepackage{url}
\usepackage[hidelinks]{hyperref}
\usepackage[utf8]{inputenc}
\usepackage[small]{caption}
\usepackage{graphicx}
\usepackage{amsmath}
\usepackage{amsthm}
\usepackage{booktabs}
\usepackage{algorithm}
\usepackage{algorithmic}
\usepackage[switch]{lineno}

\usepackage{color}

\title{A Survey on Extractive Knowledge Graph Summarization:\\Applications, Approaches, Evaluation, and Future Directions}

\author{
Xiaxia Wang$^2$
\and
Gong Cheng$^1$\thanks{Corresponding author}\\
\affiliations
$^1$State Key Laboratory for Novel Software Technology, Nanjing University, China\\
$^2$Department of Computer Science, University of Oxford, UK\\
\emails
xiaxia.wang@cs.ox.ac.uk,
gcheng@nju.edu.cn
}

\begin{document}

\maketitle

\begin{abstract}
With the continuous growth of large Knowledge Graphs (KGs), extractive KG summarization becomes a trending task. Aiming at distilling a compact subgraph with condensed information, it facilitates various downstream KG-based tasks. In this survey paper, we are among the first to provide a systematic overview of its applications and define a taxonomy for existing methods from its interdisciplinary studies. Future directions are also laid out based on our extensive and comparative review.
\end{abstract}

\section{Introduction}

\begin{figure}[t]
\centering
\includegraphics[width=\columnwidth]{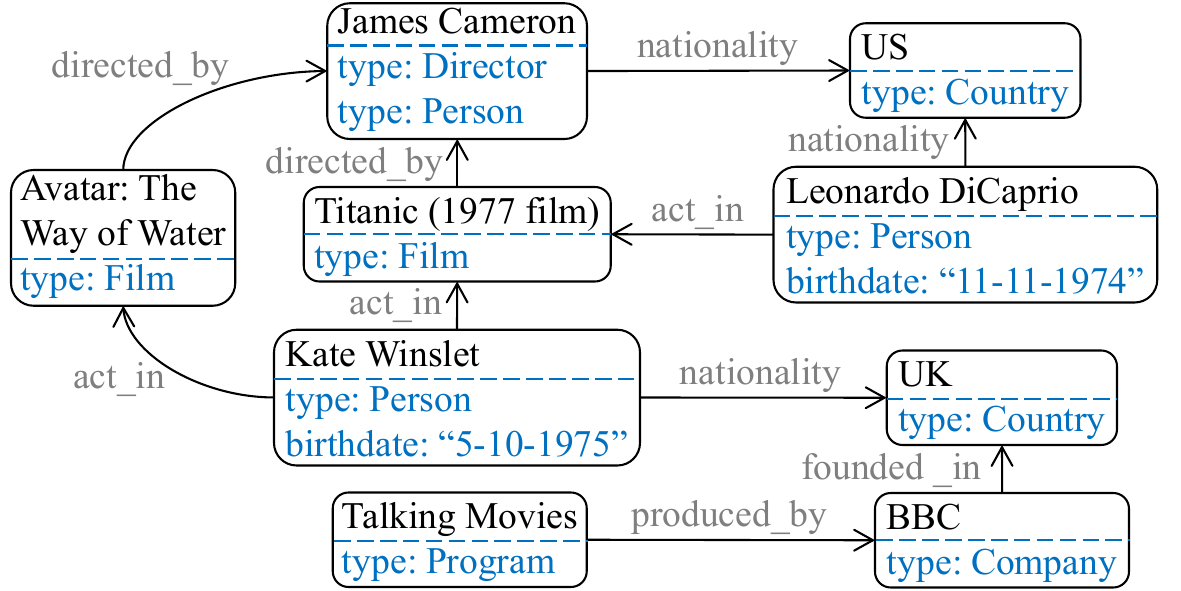}
\caption{A small KG as our running example.}
\label{fig:example}
\end{figure}

Knowledge Graph~(KG) has been a popular knowledge representation~\cite{DBLP:journals/csur/HoganBCdMGKGNNN21}. In this graph data model, as illustrated in Figure~\ref{fig:example}, nodes represent entities of interest, which are often annotated with types and attributes, and edges represent typed relations between entities; attributes and relations are collectively called properties. KGs have been widely employed in various AI systems and application fields~\cite{DBLP:journals/air/PengXNO23}, such as question answering~\cite{DBLP:conf/aaai/YangZWYW17} and machine translation~\cite{DBLP:conf/ijcai/ZhaoZZZ20}. With the emergence of large language models (LLMs), KGs have exhibited increased importance as they can also be used for training and augmenting LLMs~\cite{DBLP:journals/tgdk/PanRKSCDJO0LBMB23}.

\paragraph{Problem Description}
Given numerous KGs available on the Web, the effective discovery and selection of suitable KGs for reuse has become a challenge for humans since KGs are often large and cover a variety of topics. Such magnitude and diversity also hinder machines from efficient processing in AI systems. To tackle this problem, a straightforward idea is to generate a succinct description for a given KG to indicate its coverage or reflect its main content.
Such a size-reduced representation containing concise, meaningful, and faithfully represented information from the original KG is referred to as a \emph{summary} of the KG~\cite{DBLP:journals/vldb/CebiricGKKMTZ19}.

\paragraph{Scope of the Survey}
KG summarization is distinguished from summarizing conventional homogeneous graphs such as the Web graph and social networks~\cite{DBLP:journals/csur/LiuSDK18} by the different types of entities and relations in a KG. Accordingly, grouping-based methods for KG summarization merge entities having similar types and properties into super-nodes connected by super-edges representing relations between entities in the super-nodes. Such methods have been covered in previous surveys~\cite{DBLP:journals/vldb/CebiricGKKMTZ19,DBLP:journals/tgdk/ScherpRBCR23}. In parallel, extractive methods select an optimal subgraph from a given KG as its \emph{extractive summary}. This type of KG summary, a.k.a.~KG snippet, faithfully exemplifies the content of the original KG, and can be effortlessly comprehended by humans and directly processed by machines in the same manner as the original KG. For these advantages, extractive KG summarization has attracted considerable research attention from interdisciplinary areas related to AI and big data, and has been used in KG profiling~\cite{DBLP:journals/semweb/EllefiBBDDST18}, query optimization~\cite{DBLP:journals/semweb/HelingA23}, search~\cite{DBLP:journals/vldb/ChapmanSKKIKG20}, exploration~\cite{DBLP:conf/cidr/LissandriniMHP22}, and many other KG-driven applications. However, to the best of our knowledge, \emph{this is the first survey paper on extractive KG summarization}, providing a comprehensive overview of its applications, approaches, evaluation, and future research directions.

\begin{figure}[t]
\centering
\includegraphics[width=\columnwidth]{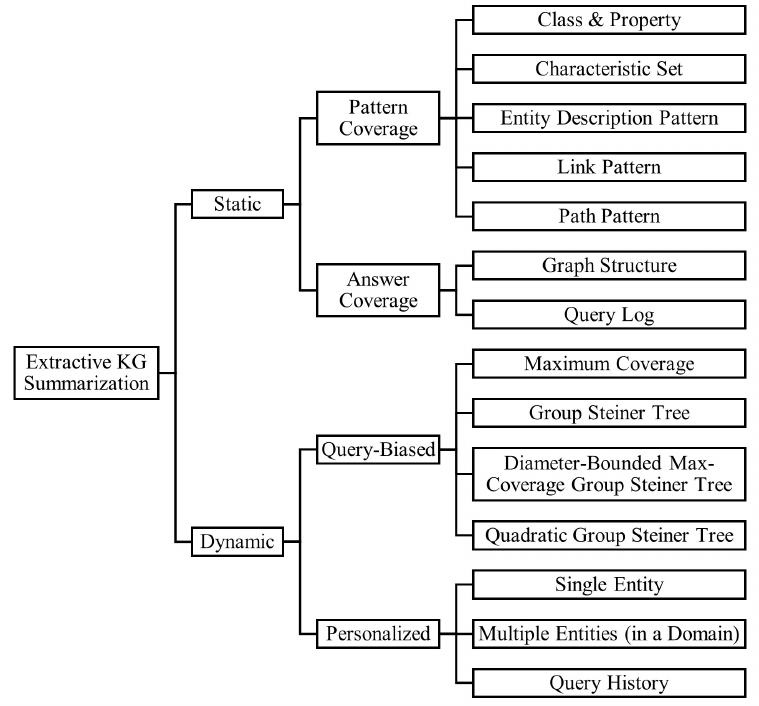}
\caption{A taxonomy of methods for extractive KG summarization.}
\label{fig:taxonomy}
\end{figure}

\paragraph{Outline}
We broadly divide extractive KG summaries into two categories: \emph{static summaries} are context-independent and always the same for a given KG; \emph{dynamic summaries} are customized according to the user's information needs. They are generated for different applications (Section~\ref{sec:applications}), by different methods (Sections~\ref{sec:approaches-static} and~\ref{sec:approaches-dynamic}), and are evaluated in different ways (Section~\ref{sec:benchmark}). We present a taxonomy of existing methods in Figure~\ref{fig:taxonomy}. To demarcate this survey, we firstly distinguish extractive KG summarization from related tasks (Section~\ref{sec:rw}).
\section{Related Work}
\label{sec:rw}

Our work differs from the following surveys on research problems that are related to extractive KG summarization.

Liu \emph{et~al.}~\shortcite{DBLP:journals/csur/LiuSDK18} reviewed a wide range of methods for \textbf{graph summarization}, but they focused mostly on homogeneous graphs rather than KGs where nodes and edges may be labeled with different types. For labeled graphs, their survey covered \textbf{grouping-based} but not extractive methods, which represent two distinct paradigms of summarization. Previous surveys on KG summarization~\cite{DBLP:journals/vldb/CebiricGKKMTZ19,DBLP:journals/tgdk/ScherpRBCR23} also gave primary attention to grouping-based methods, based on the concept of quotients in particular. Since a grouping-based summary can be viewed as an implicit schema discovered from the KG, the task is also referred to as \textbf{schema discovery}~\cite{DBLP:journals/vldb/Kellou-MenouerK22}.

\textbf{Entity summarization} selects a subset of properties for each given entity in a KG~\cite{DBLP:journals/ws/LiuCGQ21}. An entity summary can be viewed as a special type of extractive KG summary that is restricted to the neighborhood of a particular node. Such locality distinguishes entity summarization from the methods for extractive KG summarization covered in this survey which globally take the entire KG into consideration.

\textbf{Ontology summarization} extracts core concepts or a sub-ontology from a given ontology. Existing methods mainly represent an ontology as a graph to be summarized~\cite{DBLP:journals/ijsc/PouriyehALCAAMK19}. Some of these graph representations resemble a KG but are at the schema level where nodes are classes instead of entities. Accordingly, the optimization objective of ontology summarization differs significantly from those of extractive KG summarization reviewed in this survey.
\section{Applications of Extractive KG Summaries}
\label{sec:applications}

Extractive KG summarization has proved to be useful for a variety of KG-based applications. Static and dynamic KG summaries are needed for supporting different tasks.

\subsection{Applications of Static KG Summaries}

Static summaries are context-independent and are extracted to capture the intrinsic characteristics of a KG, e.g.,~to reflect its main content, or to indicate its coverage. They have found application in KG profiling and query optimization.

\subsubsection{KG Profiling}
A KG profile is a formal representation of a set of features of a given KG, which is often used to aid KG discovery, recommendation, and comparison~\cite{DBLP:journals/semweb/EllefiBBDDST18}. An extractive KG summary that contains representative entities in the original KG is one such feature understood as a sample that accurately portrays the whole KG. It has been used in KG search engines to exemplify the content of a KG getting clicked on in the search results~\cite{DBLP:journals/dint/WangLLCQ22}, aiding the user in quickly filtering relevant KGs without expensively accessing the original, potentially large KGs.

\subsubsection{KG Query Optimization}
An extractive KG summary can be regarded as a view of a given KG and queried in the same manner~\cite{DBLP:conf/icde/FanWW14,DBLP:conf/dexa/Wang17,DBLP:conf/cikm/LiCL16}. A common application of such views is federated KG query optimization~\cite{DBLP:conf/semweb/MontoyaSH17} where multiple KGs are jointly queried, and the performance relies on the efficiency of the query plan, e.g.,~the estimation of join cardinalities for sub-queries. Extractive KG summaries have been used to compute accurate estimations of this kind using less time than the complete statistics derived from the original large KGs~\cite{DBLP:journals/semweb/HelingA23}.

\subsection{Applications of Dynamic KG Summaries}

Dynamic summaries are customized and are extracted to satisfy each user’s individual information needs, e.g.,~to show the relevance of a KG to a user's query, or to tailor the content of a KG to users' specific interests.

\subsubsection{KG Search}
KG search engine has become an important tool for finding open KGs for reuse~\cite{DBLP:journals/vldb/ChapmanSKKIKG20}. Whereas early implementations rely on metadata descriptions to discover KGs and assist user judgment of their relevance, metadata has proved to be insufficient due to its quality and availability~\cite{DBLP:conf/chi/KoestenKTS17}. Complementary to metadata, various types of KG summaries have been extracted to enhance search: summaries that are centered on keyword matches to help judge the relevance of search results~\cite{DBLP:conf/cikm/ChenWCKQ19,DBLP:journals/dint/WangLLCQ22}, and compact summaries that can be fed into a dense re-ranking model (which cannot accept a whole large KG) to improve search accuracy~\cite{DBLP:conf/semweb/ChenHZLLSC23}.

\subsubsection{KG Exploration}
KG exploration refers to the process of gradually discovers and understands the content of a large and unfamiliar KG, often starting by means of a keyword query asking for matched nodes, edges, or subgraphs~\cite{DBLP:conf/cidr/LissandriniMHP22}. The matched relevant insights, as an extractive KG summary, can assist in a number of downstream tasks on this KG, such as the identification of its portions that can satisfy the current information needs, the comprehension of its graph structure, and the formulation of formal (e.g., SPARQL) queries on it.

\subsubsection{KG Reuse}
KGs such as Wikidata~\cite{DBLP:journals/cacm/VrandecicK14} are huge and cover a wide variety of topics. For a specific application, it is often sufficient to only extract a relevant portion to be reused. For example, a personalized KG summary containing the subgraph most relevant to an individual user's interests has been extracted so that it can be stored and accessed on resource-constrained devices~\cite{DBLP:conf/icdm/SafaviBFMMK19,DBLP:conf/esws/VassiliouAPK23}. Minimal domain-specific KG summaries have been extracted to reduce the computation without compromising the accuracy of domain-specific applications~\cite{DBLP:conf/bigdataconf/LalithsenaKS16,DBLP:conf/bigdataconf/LalithsenaPKS17}. Another interesting application is to enrich a news article with a relevant KG summary extracted from a background KG to help the reader comprehend the relations between entities mentioned in the news~\cite{DBLP:conf/cikm/HuangLCKQ19,DBLP:conf/ijcai/LiH0KG20}.
\section{Extraction of Static KG Summaries}
\label{sec:approaches-static}

Static summaries are extracted to reflect the coverage of the original KG. Depending on the application, existing methods cover different targets: data patterns, or query answers.

\subsection{Pattern Coverage-Based Summarization}
\label{sec:approaches-static-pattern}

\begin{figure}[h]
\centering
\includegraphics[width=\columnwidth]{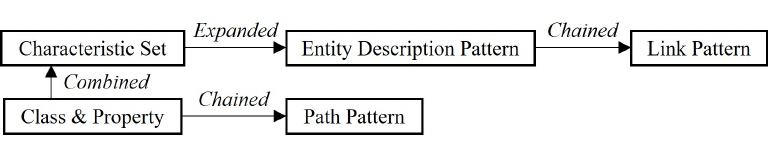}
\caption{Expressivity increases among data patterns.}
\label{fig:outline-pattern}
\end{figure}

Entities in KGs are assigned different types, i.e.,~classes, and they are described using different properties. Extracting a summary that exemplifies the most representative data patterns for describing the entities in a KG is an important component of KG profiling~\cite{DBLP:journals/semweb/EllefiBBDDST18}. Existing methods in this category have been focused on extracting summaries that cover data patterns at different granularities, from independent instantiations of classes and properties to their joint instantiations associated with graph structure. Figure~\ref{fig:outline-pattern} outlines the expressivity increases among different data patterns.

\subsubsection{Class \& Property}
\emph{Class and property} instantiations are the most elementary data patterns in a KG, such as the instances of the class \texttt{Person} and the property \texttt{nationality} in Figure~\ref{fig:example}. To exemplify all such instantiations during KG visualization, the 3-S approach~\cite{DBLP:conf/icde/SundaraAKDWCS10} performs stratified sampling by uniformly sampling the instances of each class and each property independently.
However, this method is only suitable for KGs that have a small schema.

\begin{figure}[h]
\centering
\includegraphics[width=\columnwidth]{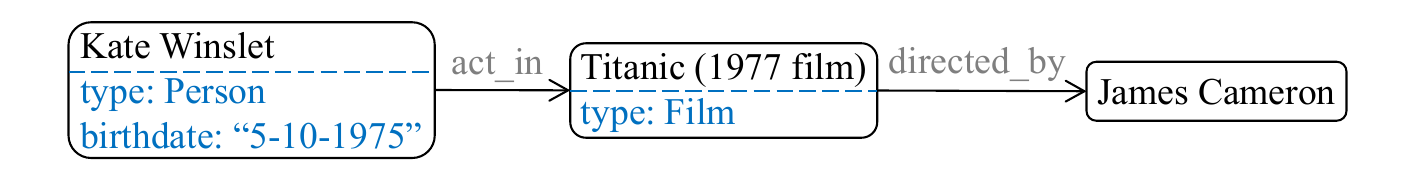}
\caption{A summary extracted from Figure~\ref{fig:example} containing the most frequently instantiated class \texttt{Person} and property \texttt{act\_in}.}
\label{fig:illusnip}
\end{figure}

Given a potentially large number of classes and properties used in a KG, the IlluSnip approach~\cite{DBLP:conf/wsdm/ChengJDXQ17} extracts a size-bounded connected subgraph that contains the most frequently instantiated classes and properties, as well as entities having the highest PageRank scores, to illustrate the main content of the KG. For example, the summary in Figure~\ref{fig:illusnip} includes \texttt{Person} and \texttt{act\_in} which are the most frequently instantiated class and property in Figure~\ref{fig:example}, respectively. To fulfill this, a \emph{Maximum-weight-and-coverage Connected Graph} problem (MwcCG) is formulated, and is solved by a greedy approximation algorithm for this new NP-hard optimization problem, where coverage and weights are jointly maximized to account for class/property instantiations and entity scores, respectively.

As a follow-up effort, Liu et al.~\shortcite{DBLP:journals/tweb/LiuCLQ19} design more efficient algorithms for MwcCG and adapt them to SPARQL endpoints for summarizing remotely accessed KGs.

\subsubsection{Characteristic Set}
On top of individual properties, the \emph{characteristic set}~\cite{DBLP:conf/esws/HelingA20} of an entity is the set of all properties (including \texttt{type}) describing that entity in a KG, and each property can be associated with its average frequency of occurrence among all the entities having this characteristic set. For example, the characteristic set of \texttt{James Cameron} in Figure~\ref{fig:example} consists of $\{$\texttt{nationality}, \texttt{type}$\}$. All the characteristic sets instantiated in a KG, the number of entities having each characteristic set, and the average frequency of each property constitute a statistical profile of the KG which is required by a query optimizer for devising efficient query plans. When it is difficult to access or process the entire KG, Heling et al.~\shortcite{DBLP:conf/esws/HelingA20,DBLP:journals/semweb/HelingA23} use estimations for this profile derived from a KG summary containing the properties of a set of selected entities. Entities that have high outdegrees (i.e.,~rich descriptions) have a high chance of being selected.

\subsubsection{Entity Description Pattern}
\begin{figure}[h]
\centering
\includegraphics[width=\columnwidth]{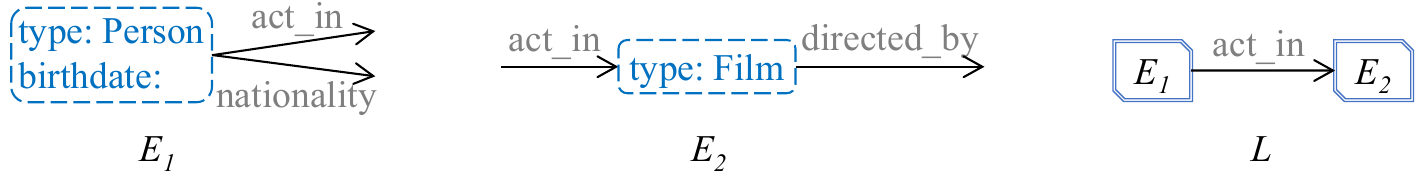}
\caption{Two EDPs ($E_1, E_2$) and a LP ($L$) in Figure~\ref{fig:example}.}
\label{fig:edp-lp}
\end{figure}

In addition to the properties (i.e.,~outgoing edges) in a characteristic set, the \emph{Entity Description Pattern} (EDP)~\cite{DBLP:conf/semweb/00010LXPKQ21} of an entity distinguishes its types from other properties and, further, includes the properties that have this entity as a value (i.e.,~incoming edges).
For example, Figure~\ref{fig:edp-lp} shows~$E_1$, the EDP of two entities \texttt{Kate Winslet} and \texttt{Leonardo DiCaprio} in Figure~\ref{fig:example}, and~$E_2$, the EDP of \texttt{Titanic (1977 film)} and \texttt{Avatar: The Way of Water}.
To extract a summary of a KG to exemplify all or the most frequent EDPs, the two-step PCSG approach~\cite{DBLP:conf/semweb/00010LXPKQ21}, in its first step, selects the minimum number of connected components that collectively cover those EDPs by formulating and solving a \emph{Set Cover} problem. In the second step, it extracts an optimal subgraph from each selected connected component and merges these subgraphs. Specifically, the subgraph is a tree found by solving a minimum-size \emph{Group Steiner Tree} problem~\cite{DBLP:conf/wg/Ihler91} where all the entities having the same EDP form a group. The extracted minimal tree is required to cover all the groups, i.e., contain at least one entity for each EDP; the types and properties of this entity are then extracted to exemplify its EDP.

\subsubsection{Link Pattern}
\begin{figure}[h]
\centering
\includegraphics[width=\columnwidth]{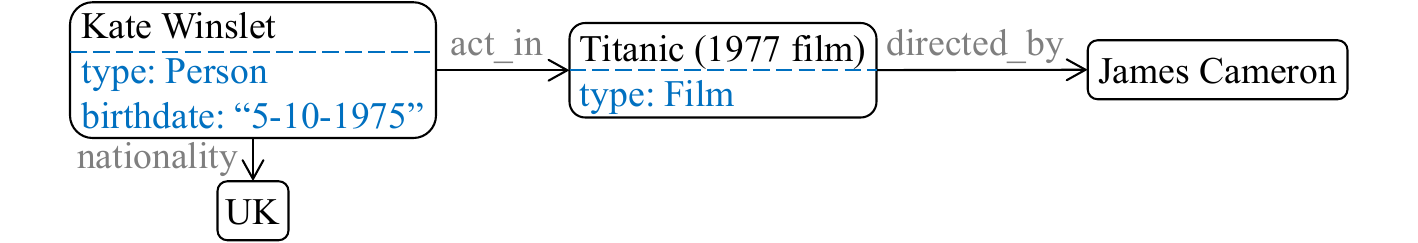}
\caption{A summary extracted from Figure~\ref{fig:example} exemplifying the two EDPs and the LP in Figure~\ref{fig:edp-lp}.}
\label{fig:pcsg}
\end{figure}

Building on EDPs, \emph{Link Pattern} (LP)~\cite{DBLP:conf/semweb/00010LXPKQ21} takes a further step to characterize not only the neighborhood pattern of each individual entity, but also the patterns of relations between entities. Specifically, the LP of a relation between two entities is a triple consisting of the relation type and the EDPs of the two entities. For example, Figure~\ref{fig:edp-lp} shows a LP shared by the three \texttt{act\_in} relations in Figure~\ref{fig:example}.
To extract a KG summary that also exemplifies all or the most frequent LPs, the aforementioned PCSG approach~\cite{DBLP:conf/semweb/00010LXPKQ21} extends the formulation of the Group Steiner Tree problem by performing edge subdivision to convert each relation into a node; all the relations having the same LP form a group to be covered by the extracted minimal tree. For example, the summary in Figure~\ref{fig:pcsg} exemplifies the two EDPs and the LP in Figure~\ref{fig:edp-lp}. Note that the EDPs of the entities \texttt{James Cameron} and \texttt{UK} are not fully exemplified in this summary.

\subsubsection{Path Pattern}
\begin{figure}[h]
\centering
\includegraphics[width=\columnwidth]{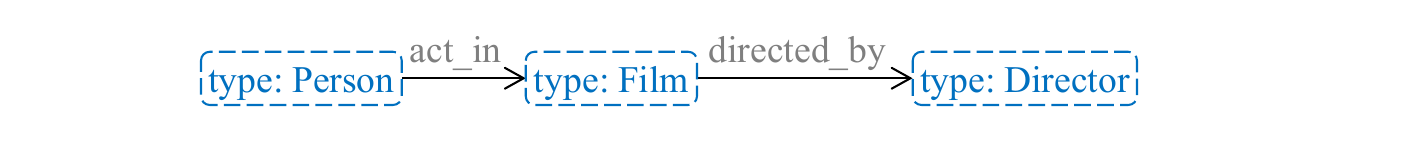}
\caption{A path pattern in Figure~\ref{fig:example}.}
\label{fig:path-pattern}
\end{figure}

Going beyond link patterns that focus on the direct connection patterns between adjacent entities, the more generalized \emph{path pattern}~\cite{DBLP:conf/i-semantics/MynarzDTS16} addresses both direct and indirect connections between entities. It is an alternating sequence of classes and properties, derived from the paths in a KG where each entity is replaced by a type assigned to it. For example, Figure~\ref{fig:path-pattern} shows the most frequent path pattern of length two in Figure~\ref{fig:example}. To present typical patterns for user comprehension, for each frequent path pattern in a KG, Mynarz et al.~\shortcite{DBLP:conf/i-semantics/MynarzDTS16} extract a set of diverse and representative instance paths. Diversification is achieved by choosing entities described using different properties, and representative entities are found by clustering entities having similar properties and then choosing from the medoids of the clusters.

\paragraph{Remarks}
An extractive summary aiming at covering data patterns at a low granularity such as class and property instantiations is able to concisely exemplify the content of a KG. However, higher-order structural patterns like EDPs provide more accurate characterization, accompanied with increased computational complexity and also redundancy, e.g., two EDPs may differ insignificantly by only one property. This trade-off is to be considered by the concrete application.

\subsection{Answer Coverage-Based Summarization}

Replacing a large KG with an extracted compact summary to be queried helps improve query performance~\cite{DBLP:conf/icde/FanWW14}. Existing methods in this category are targeted at maximizing a summary's coverage of query answers~\cite{DBLP:conf/semweb/RietveldHSG14} or its preservation of statistic features of data distribution in the original KG~\cite{DBLP:conf/semweb/MontoyaSH17}. The latter is important to federated query optimization, for which some methods reviewed in Section~\ref{sec:approaches-static-pattern} perform a biased sampling of entities according to their outdegrees~\cite{DBLP:conf/esws/HelingA20,DBLP:journals/semweb/HelingA23}, which is a common measure of node centrality in graphs. Below we will see other centrality measures and the utilization of query logs.

\subsubsection{Graph Structure}
Without relying on prior knowledge of queries, the SampLD approach~\cite{DBLP:conf/semweb/RietveldHSG14} attempts to purely employ graph topology to predict the relevance of edges for typical queries. Assuming that structurally central entities are likely to be included in common query answers, this approach extracts edges that are incident with the most central nodes in a KG. Three graph centrality measures are applied and compared: \emph{PageRank}, \emph{indegree}, and \emph{outdegree}. For example, if indegree is used, the edge between \texttt{James Cameron} and \texttt{Titanic (1977 film)} in Figure~\ref{fig:example} will be ranked high since these two nodes have the largest indegree in the KG.

\subsubsection{Query Log}
Besides graph centrality, it would be reasonable to assume that entities and properties that occur frequently in past queries or in the results of past queries are important and are likely to be included in the answers of future queries. For example, if the entity \texttt{BBC} in Figure~\ref{fig:example} is frequently queried according to the query log, it will have a high priority to be included in the summary despite its relatively small indegree and outdegree in the KG. Guo et al.~\shortcite{DBLP:conf/service/GuoW21} incorporate this idea and propose a hybrid approach that takes into account both graph structure and query log. They measure the importance of an edge by a combination of graph centrality (i.e., the indegree and outdegree of its endpoints) and query frequency (i.e., the frequency of co-occurrence of its endpoints in the query log). Node importance is measured in an analogous manner.
Finally. top-ranked edges that are incident with top-ranked nodes are extracted to form a summary.

\paragraph{Remarks}
Graph structure and query log are complementary to each other. Whereas observed query frequency represents an effective indicator for the likelihood of being queried in the future, when query log is not available, graph centrality could be used as a reasonable estimation for substitution.
\section{Extraction of Dynamic KG Summaries}
\label{sec:approaches-dynamic}

Dynamic summaries are extracted to tailor a KG to users' needs. Existing methods consider needs expressed in different manners: keyword queries, or personal interests.

\subsection{Query-Biased Summarization}

\begin{figure}[h]
\centering
\includegraphics[width=\columnwidth]{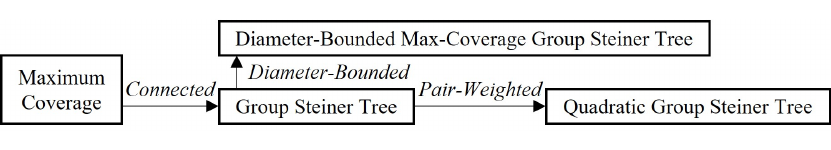}
\caption{Expressivity increases among problem formulations for query-biased extractive KG summarization.}
\label{fig:outline-gst}
\end{figure}

Snippet generation for KG search~\cite{DBLP:conf/cikm/ChenWCKQ19} and query-based KG exploration~\cite{DBLP:conf/cidr/LissandriniMHP22} can both be formulated as extracting a query-biased subgraph from a given KG. Early methods in this category solve it as a node/edge ranking problem. Recently, various formulations in combinatorial optimization have been adopted, in particular based on Group Steiner Trees for reflecting connections among query keywords in the graph structure. Not limited to maximizing the coverage of query keywords, they consider answer compactness and cohesiveness. Figure~\ref{fig:outline-gst} outlines the expressivity increases among different problem formulations.

\subsubsection{Early Heuristics}
To acquaint the user with a KG in the context of keyword search, in an early work, Bai et al.~\shortcite{DBLP:conf/otm/BaiDT08} extract two subgraphs: a \emph{topic-oriented subgraph} for revealing the main topic of the KG, and a \emph{query-oriented subgraph} for showing its relevance to the query.
The topic-oriented subgraph is extracted from the neighborhood of a topic entity based on predefined priorities of its properties; the topic entity is one that has a large indegree and outdegree. The query-oriented subgraph contain edges matched with query keywords; priority goes first to the edges incident with the topic entity.

\subsubsection{Maximum Coverage}
\begin{figure}[h]
\centering
\includegraphics[width=\columnwidth]{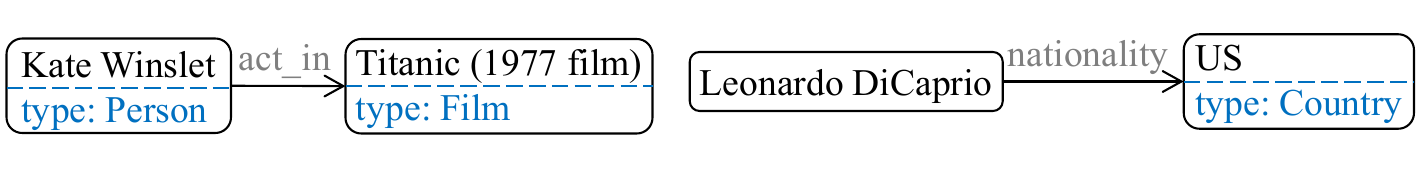}
\caption{A summary extracted from Figure~\ref{fig:example} containing the most frequently instantiated class \texttt{Person} and property \texttt{act\_in}, and containing the query keywords \textit{Titanic} and \textit{US}.}
\label{fig:ksd}
\end{figure}

Instead of separately considering data representativeness and query relevance in two subgraphs, the KSD approach~\cite{DBLP:conf/semweb/WangCK19} jointly maximizes the coverage of all such elements of interest. Specifically, the elements of a KG that a summary is expected to cover include class and property instantiations, central entities, and keyword matches. To achieve it, a weighted \emph{Maximum Coverage} problem is formulated, where each edge (or type/attribute assertion) is represented as a set that may cover a class instantiation, a property instantiation, one or two entities, and/or a number of query keywords. Each class and property to be covered is weighted by its frequency in the KG. Each entity to be covered is assigned a weight measuring its graph centrality based on indegree and outdegree.
Solving this problem gives rise to a size-bounded optimum subgraph that maximizes the coverage of representative and query-biased data elements of the KG. As illustrated in Figure~\ref{fig:ksd}, the summary extracted by this approach is comparable with the summary extracted by IlluuSnip shown in Figure~\ref{fig:illusnip}. Given a keyword query \textit{Titanic US}, whereas both summaries include the most frequently instantiated class \texttt{Person} and property \texttt{act\_in} in Figure~\ref{fig:example}, the summary here further includes the entity \texttt{US} to match the query, although it is not required to be a connected subgraph.

\subsubsection{Group Steiner Tree}
\begin{figure}[h]
\centering
\includegraphics[width=\columnwidth]{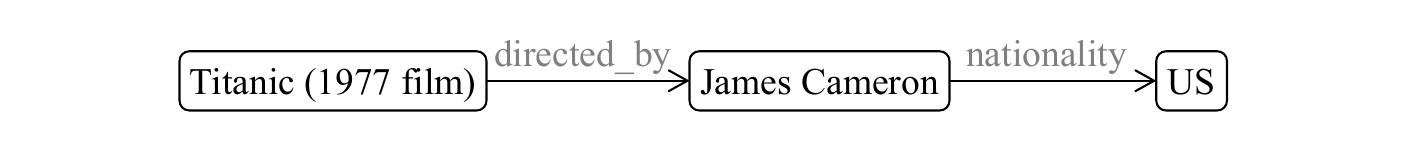}
\caption{A summary extracted from Figure~\ref{fig:example} containing a path that connects the query keywords \textit{Titanic} and \textit{US}.}
\label{fig:gst}
\end{figure}

Independent coverage of KG elements will easily produce a summary composed of multiple disconnected fragments. This weakens the structural cohesiveness of the summary, and misses the connections between keyword matches in the KG, which could be important to the user's judgment of query relevance. To overcome this limitation, a recent line of research formulates and solves a minimum-weight \emph{Group Steiner Tree} (GST) problem~\cite{DBLP:conf/wg/Ihler91}. The KeyKG approach~\cite{DBLP:conf/www/Shi0K20} maps each query keyword to a group of entities, and the target is a minimum-weight tree that covers all the groups, i.e., contains at least one matched entity for each keyword. Edge weights can be defined by an off-the-shelf method depending on the application. The resulting tree concisely reflects a structural relationship among all the query keywords. For example, given a keyword query \textit{Titanic US}, the summary in Figure~\ref{fig:gst} includes a path that connects the two query keywords. Considering the NP-hardness of this problem, KeyKG features a new approximation algorithm that employs an offline computed distance index to accelerate online extraction of shortest paths between groups to be merged into a tree.

\begin{figure}[h]
\centering
\includegraphics[width=\columnwidth]{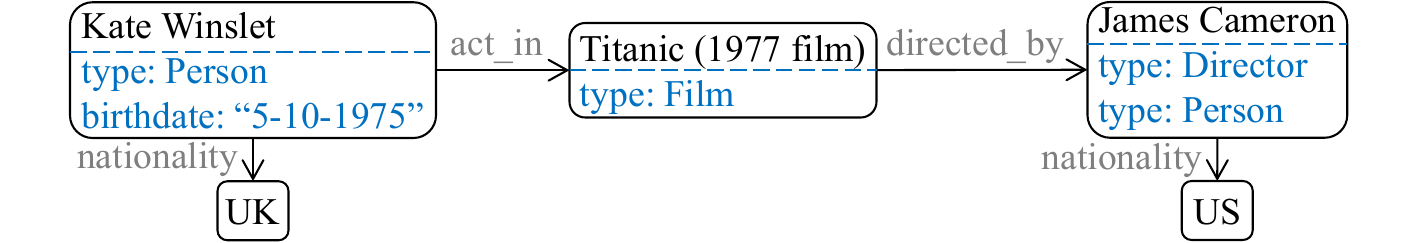}
\caption{A summary extracted from Figure~\ref{fig:example} exemplifying the two EDPs and the LP in Figure~\ref{fig:edp-lp}, and containing the query keywords \textit{Titanic} and \textit{US}.}
\label{fig:qpcsg}
\end{figure}

The aforementioned PCSG approach incorporates KeyKG as an efficient solver of the GST problem. Moreover, PCSG has been extended to QPCSG~\cite{DBLP:conf/semweb/00010LXPKQ21}, which covers EDPs, LPs, and also query keywords. All the entities and relations matched with a query keyword form a group to be covered. As illustrated in Figure~\ref{fig:qpcsg}, given a keyword query \textit{Titanic US}, the summary extracted by this extended approach further includes the entity \texttt{US} to match the query, extending the summary in Figure~\ref{fig:pcsg} extracted by PCSG.

\subsubsection{Diameter-Bounded Max-Coverage Group Steiner Tree}
The standard GST problem requires the extracted tree to cover all the groups, but this is not always achievable in practice. For example, given a KG that is a disconnected graph, a tree that connects all the keywords in a query may not exist. Even for a connected graph, a large tree may have to be extracted to cover keywords that are distant from each other in the graph structure, against the goal of extracting a compact summary. To achieve a trade-off between query coverage and structural compactness, Cheng et al.~\shortcite{DBLP:conf/semweb/ChengLZL20} propose to allow relaxing a query by ignoring a minimum number of keywords to guarantee the compactness of the extracted summary.

\begin{figure}[h]
\centering
\includegraphics[width=\columnwidth]{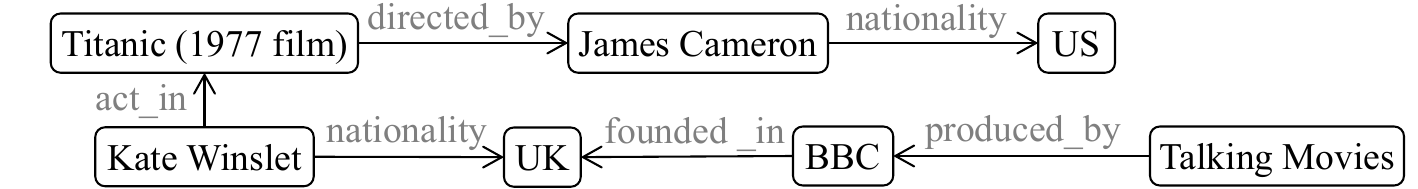}
\caption{A summary extracted from Figure~\ref{fig:example} containing a long path that connects the query keywords \textit{Titanic}, \textit{US}, and \textit{talking}.}
\label{fig:non-dcgst}
\end{figure}

As a generalization of this idea, Zhang et al.~\shortcite{DBLP:conf/www/ZhangW023} formulate a \emph{Diameter-bounded max-Coverage Group Steiner Tree} (DCGST) problem. Compared with a standard GST, DCGST has a bounded diameter and is required to cover not necessarily all but the most possible groups. For example, given a keyword query \textit{Titanic US talking} over Figure~\ref{fig:example} and a diameter bound of two hops, the summary extracted here has to ignore the keyword \textit{talking} but choose to be concisely focused on the connection between the other two keywords, as illustrated in Figure~\ref{fig:gst}. By contrast, the summary extracted by the aforementioned KeyKG approach is a standard GST consisting of a long and unfocused path, as shown in Figure~\ref{fig:non-dcgst}, although it covers all the three keywords. The DCGST problem is an emerging NP-hard optimization problem and is solved by PrunedCBA~\cite{DBLP:conf/www/ZhangW023}, an approximation algorithm performing pruned best-first search based on a novel hop-bounded index of precomputed distances.

\subsubsection{Quadratic Group Steiner Tree}
\begin{figure}[h]
\centering
\includegraphics[width=\columnwidth]{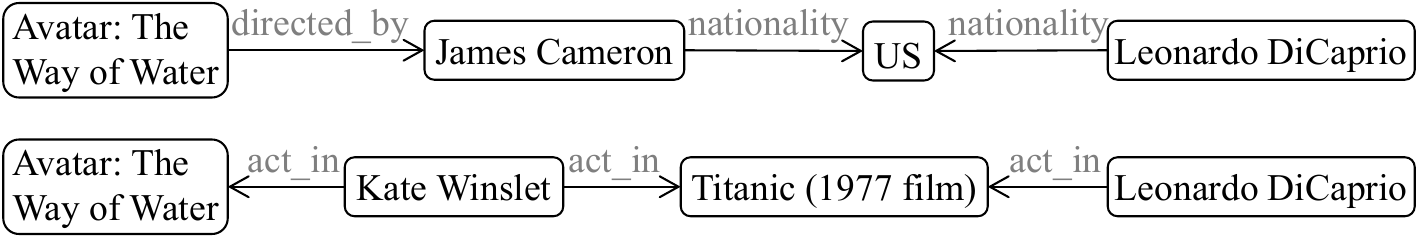}
\caption{Two summaries extracted from Figure~\ref{fig:example} containing the query keywords \textit{Avatar} and \textit{DiCaprio}.}
\label{fig:qgst}
\end{figure}

Another extension of the GST formulation for query-biased KG summarization considers the semantic cohesiveness of the extracted tree~\cite{DBLP:conf/bigdataconf/ChengK17}. Recall that the standard minimum-weight GST problem optimizes the total weight of the extracted nodes or edges, where weights reflect (inverse) salience. However, a set of salient nodes or edges may not always comprise a meaningful connection as a whole. It can be a semantically disjointed tree that connects a set of salient but disparate entities. To improve semantic cohesiveness, Shi et al.~\shortcite{DBLP:conf/www/Shi0TKS21,DBLP:conf/ijcai/Shi0T0K21} incorporate the minimization of semantic distances between extracted entities into the objective function. They formulate a more generalized \emph{Quadratic Group Steiner Tree} (QGST) problem. Compared with a standard GST, QGST minimizes a linear combination of the sum of weights of its nodes and the sum of quadratic weights of its node pairs. Quadratic weight characterizes the semantic distance between a pair of entities, e.g., the angular distance between their embedding vectors. The extracted summary tends to include entities that are not only salient by themselves and matched with the query, but also semantically close to each other, hence forming a semantically cohesive whole. For example, given a keyword query \textit{Avatar DiCaprio}, the top tree in Figure~\ref{fig:qgst} contains salient entities like \texttt{US} but also disparate entities including a film, a director, a country, and an actor. By contrast, the entities in the bottom tree are a set of closely related films and actors, representing a more meaningful connection. To solve the QGST problem which is an NP-hard optimization problem, B$^3$F~\cite{DBLP:conf/ijcai/Shi0T0K21} is an exact algorithm that performs branch-and-bound best-first search. QO and EO~\cite{DBLP:conf/www/Shi0TKS21} are two approximation algorithms with the idea of finding and merging small-weight paths from a root node to all the groups.

\paragraph{Remarks}
Compared with maximum coverage, GST-based formulations are expressive but hard to solve. Although KeyKG and PrunedCBA perform satisfyingly on large KGs with millions of nodes, they rely on a precomputed distance index which needs to be rebuilt when the KG evolves. For the QGST problem, no algorithm so far could respond to a query over a million-scale KG in real time. All these limitations should be taken into account when choosing the formulation.

\subsection{Personalized Summarization}

Besides keyword queries, users' interests can be explicitly expressed or implicitly discovered in various manners. Serving the user with a personalized summary extracted from a large KG provides cost-effectiveness in KG reuse~\cite{DBLP:conf/icdm/SafaviBFMMK19,DBLP:conf/esws/VassiliouAPK23}. Existing methods in this category tailor KGs to a user's needs in different forms: a single entity of interest, a class representing a domain or a set of entities specified by the user, or the user's own query history.

\subsubsection{Single Entity}
The iSummary approach~\cite{DBLP:conf/esws/VassiliouAPK23} allows the user to specify an entity of interest in the KG as a \emph{seed entity}, such as \texttt{James Cameron}, and it extracts a summary that compactly describes this entity as well as a set of its closely related entities. Specifically, each entity in the KG is assigned a weight. An entity with a higher frequency of co-occurrence with the seed entity in the query log has a higher weight. This frequency is believed to reflect the relatedness between entities. From the resulting node-weighted graph, a maximum-weight tree that connects the seed entity with its most related entities is extracted. This computational problem resembles the minimum-weight \emph{Steiner Tree} problem and is NP-hard. It is solved by an approximation algorithm that extracts shortest paths to connect the seed entity with its most related entities.

\subsubsection{Multiple Entities (in a Domain)}
A user may also want to extract a domain-specific summary from a large KG. A domain of interest can be represented by a set of \emph{in-domain entities}, e.g., all the entities having a particular type such as \texttt{Film} in Figure~\ref{fig:example}. To fulfill this need, Lalithsena et al.~\shortcite{DBLP:conf/bigdataconf/LalithsenaKS16} traverse the KG starting from in-domain entities to extract domain-specific edges. Their primary focus is to determine the domain specificity score of each relation type, and they present two measures. The first measure utilizes the strength of association among the types of relations and intermediate entities derived from instance-level frequencies. The second and more effective measure relies on the domain specificity of intermediate relations and adopts Pointwise Mutual Information for measuring association.

In a follow-up work, Lalithsena et al.~\shortcite{DBLP:conf/bigdataconf/LalithsenaPKS17} further employ the Wikipedia categories of each entity and restrict graph traverse to the entities in the top-ranked domain-specific categories. To determine the domain specificity score of a category, they collect evidences from the types of the entities in the category, its lexical label, and its structural abstractness characterized by its outdegree in the category hierarchy. Evidences are aggregated for ranking categories using Probabilistic Soft Logic, a statistical relational learning framework.

Instead of graph traversal, Jiang et al.~\shortcite{DBLP:conf/adma/JiangZGGWZ12} firstly mine the frequent subgraph patterns in a KG, and extract their instance subgraphs that contain user-specified entities. These subgraphs are merged by intersection or union into a single graph, where each edge is weighted based on a combination of the frequencies of its constituent entities and relation type. Finally, a minimum-weight Steiner Tree that connects all the user-specified entities is extracted from this graph as a summary of the semantic associations among those entities.

Here we would like to distinguish the approaches in this category from the research on \emph{relationship search}~\cite{DBLP:journals/sigweb/000120}, i.e.,~searching and ranking subgraphs that connect a given set of input entities~\cite{DBLP:journals/tkde/ChengSQ17}.
Their main difference is that the approaches here extract and summarize all the subgraphs relevant to the input entities, whereas relationship search is only focused on finding a few top-ranked subgraphs to be presented as search results.

\subsubsection{Query History}
When a user's personal query history is available, the GLIMPSE approach~\cite{DBLP:conf/icdm/SafaviBFMMK19} extracts a personalized summary from a KG containing only a subgraph most relevant to the user's interest reflected in past queries. To this end, entities and relations are scored by a probabilistic framework estimated from their frequencies observed in the user's query history. The extraction of a size-bounded optimum subgraph is formulated as an optimization problem with a submodular objective function, hence featuring a greedy constant-factor approximation algorithm.

\paragraph{Remarks}
Despite the diverse forms of expressing a user's needs, it is possible to convert them into each other to be handled in a consistent manner. For example, from a user's query history, a set of frequent entities can be mined, which, in turn, can be processed separately as single entities.
\section{Evaluation of Extractive KG Summaries}
\label{sec:benchmark}

Considering the magnitude of a KG, it is difficult---if not impossible---to manually extract a summary as the gold standard for evaluation. Instead, depending on the application, various \emph{quality metrics} have been used to quantitatively assess the quality of a summary from different perspectives. For example, Wang et al.~\shortcite{DBLP:journals/tkde/WangCPKQ23} devise an evaluation framework for measuring a summary's coverage of the classes, properties, EDPs, and LPs in the original KG. They also publish a benchmark called BANDAR with thousands of real-world KGs to be used for evaluation.
Cheng et al.~\shortcite{DBLP:conf/wsdm/ChengJDXQ17} conduct a user study, inviting human users to rate and compare the quality of the summaries extracted by different approaches.

As to task-specific summaries, \emph{extrinsic evaluation} aims to assess the quality of a summary indirectly by testing its performance in downstream tasks using the summary. For example, Rietveld et al.~\shortcite{DBLP:conf/semweb/RietveldHSG14} and Guo et al.~\shortcite{DBLP:conf/service/GuoW21} measure a summary's coverage of query answers by comparing the results retrieved from the summary with the correct results retrieved from the original KG.
Wang et al.~\shortcite{DBLP:conf/semweb/00010LXPKQ21} conduct a user study to compare the usefulness of the summaries extracted by different approaches for assisting users in comprehending a large KG and then completing complex SPARQL queries over the KG. Useful summaries are expected to help complete SPARQL queries correctly and also quickly.

Besides, the \emph{time} for extracting a summary is an important factor in evaluating the practicability of an algorithm, especially for real-time applications such as KG search engines.

\section{Conclusion and Future Directions}
\label{sec:con}

We have systematically reviewed existing applications, approaches, and evaluation methods for extractive KG summarization. This interdisciplinary research topic spanning artificial intelligence, data mining, and information retrieval provides increasing research opportunities. We highlight the following research directions that are currently underexplored.

\paragraph{Neural Extraction}
While existing approaches mainly exploit the symbolic features of KGs (e.g., ontological schema, graph structure), there is much room for exploring neural methods, given the rapid advances in deep learning models in recent years, particularly graph neural networks.

\paragraph{Supervised Extraction}
Although unsupervised extraction has become mainstream to KG summarization, supervised methods deserve to be taken into consideration. The inadequacy of labeled data for training could be addressed by semi-supervised learning such as self-training and co-training.

\paragraph{Generative Extraction}
Extractive summaries are not necessarily extracted. Motivated by the recent success in generative information extraction and generative information retrieval, it would be an interesting investigation to apply pretrained generative models to KG summary extraction.

\paragraph{Comparative Extraction}
When multiple KGs need to be summarized and compared, e.g., in the results page of a KG search engine, the extraction of their summaries should not be independent but can be performed jointly, attending to their common and distinctive features to facilitate user's selection.

\paragraph{Collaborative Extraction}
Given the computational cost of processing a large KG, we anticipate a scenario where summaries are extracted not from scratch but building on stored summaries previously extracted for similar tasks, which can be inexpensively assembled and tailored to new needs.

\clearpage

\section*{Acknowledgments}
This work was supported by the NSFC (62072224).

\bibliographystyle{named}
\bibliography{main}

\end{document}